%% file: main.tex
\input{templates/constants}

\pdfoutput=1
\documentclass[10pt,twocolumn,letterpaper]{article}
\input{templates/header}
\begin{document}
\title{\paperTitle}
\author{\authorBlock}
\input{figures/01_Teaser}
\input{chapters/00_abstracts}
\input{chapters/01_intro}

\input{chapters/02_related}
\input{chapters/03_method}
\input{chapters/04_experiments}
\input{chapters/05_conclusion}
\input{chapters/06_acknowledgement}

{\small
\bibliographystyle{templates/ieeenat_fullname}

\input{main.bbl}
}

\clearpage \appendix \input{chapters/07_appendix}
\end{document}

%% file: templates/constants.tex
\def\paperTitle{WHAM: Reconstructing World-grounded Humans with Accurate 3D Motion}

\def\authorBlock{
    Soyong Shin$^{1}$ \qquad
    Juyong Kim$^{1}$ \qquad
    Eni Halilaj$^{1}$ \qquad
    Michael J. Black$^{2}$ \vspace{3mm} \\
    $^{1}$Carnegie Mellon University \hspace{5mm} $^{2}$Max Planck Institute for Intelligent System \\
}


%% file: templates/header.tex
\usepackage[pagenumbers]{templates/style}
\input{templates/macros}  

\usepackage{xr-hyper}
\usepackage{indentfirst}
\usepackage{cuted}
\usepackage{capt-of}
\usepackage{fontawesome}

\makeatletter
\newcommand*{\addFileDependency}[1]{
  \typeout{(#1)}
  \@addtofilelist{#1}
  \IfFileExists{#1}{}{\typeout{No file #1.}}
}

\makeatother

\definecolor{arxivblue}{rgb}{0.21,0.49,0.74}
\usepackage[pagebackref,breaklinks,colorlinks,citecolor=arxivblue]{hyperref}
\usepackage[capitalize]{cleveref}
\crefname{section}{Sec.}{Secs.}
\crefname{table}{Table}{Tables}
\crefname{figure}{Fig.}{Figs.}

%% file: templates/macros.tex
\usepackage{graphicx}	
\usepackage{amsmath}	
\usepackage{amssymb}	
\usepackage{booktabs}
\usepackage{times}
\usepackage{microtype}
\usepackage{epsfig}
\usepackage[table,xcdraw,dvipsnames]{xcolor}
\usepackage{caption}
\usepackage{float}
\usepackage{placeins}
\usepackage{color, colortbl}
\usepackage{stfloats}
\usepackage{enumitem}
\usepackage{tabularx}
\usepackage{xstring}
\usepackage{multirow}
\usepackage{xspace}
\usepackage{url}
\usepackage{subcaption}
\usepackage{xcolor}
\usepackage[hang,flushmargin]{footmisc}

\usepackage[accsupp]{axessibility}

% ...

\newcommand{\R}[1]{{%
    \textbf{%
        \ifstrequal{#1}{1}{\textcolor{red}{R#1}}{%
        \ifstrequal{#1}{2}{\textcolor{blue}{R#1}}{%
        \ifstrequal{#1}{3}{\textcolor{magenta}{R#1}}{%
        \ifstrequal{#1}{4}{\textcolor{teal}{R#1}}{%
                           \textcolor{cyan}{R#1}%
        }}}}%
    }%
}}

%% file: figures/01_Teaser.tex
\twocolumn[{%
  \renewcommand\twocolumn[1][]{#1}%
  \maketitle
  \vspace*{-1cm}
   \begin{center}
  \centerline{\includegraphics[width=1.0\textwidth]{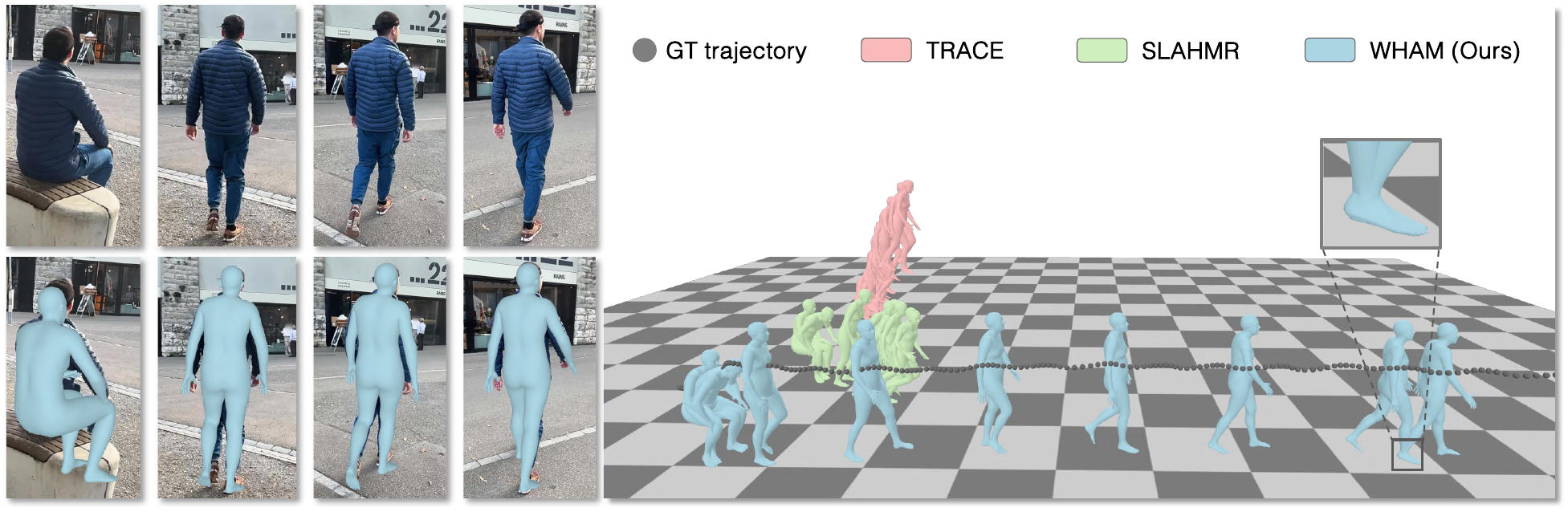}}
   \vspace*{-2.5em}
  \end{center}

\begin{center}
\captionof{figure}{\textbf{WHAM: World-grounded Humans with Accurate Motion.} 
State-of-the-art methods like \textcolor{Salmon}{TRACE} \cite{trace} and \textcolor{YellowGreen}{SLAHMR} \cite{slahmr} 
fail to capture global 3D human trajectories accurately when given in-the-wild videos captured using a moving camera, producing implausible world-grounded motion (e.g., foot sliding). 
To address this, \textcolor{Turquoise}{WHAM} uses two novel strategies: (1) feature integration from 2D keypoints and pixels to reconstruct precise and pixel-aligned 3D human motion and (2) contact-aware trajectory recovery to place the human in a global coordinate system without foot sliding. 
Gray dots show the ground-truth global trajectory.
See \textbf{Supplemental Video.}\vspace*{-0.2cm}
\label{fig:teaser}}
\end{center}
\vspace*{0.2cm}
}]

%% file: chapters/00_abstracts.tex
\begin{abstract}
    The estimation of 3D human motion from video has progressed rapidly but current methods still have several key limitations. First, most methods estimate the human in camera coordinates. Second, prior work on estimating humans in global coordinates often assumes a flat ground plane and produces foot sliding. Third, the most accurate methods rely on computationally expensive optimization pipelines, limiting their use to offline applications. Finally, existing video-based methods are surprisingly less accurate than single-frame methods. We address these limitations with WHAM (World-grounded Humans with Accurate Motion), which accurately and efficiently reconstructs 3D human motion in a global coordinate system from video. WHAM learns to lift 2D keypoint sequences to 3D using motion capture data and fuses this with video features, integrating motion context and visual information. WHAM exploits camera angular velocity estimated from a SLAM method together with human motion to estimate the body's global trajectory. We combine this with a contact-aware trajectory refinement method that lets WHAM capture human motion in diverse conditions, such as climbing stairs. WHAM  outperforms all existing 3D human motion recovery methods across multiple in-the-wild benchmarks. Code is available for research purposes at \href[pdfnewwindow=true]{http://wham.is.tue.mpg.de/}{http://wham.is.tue.mpg.de/}.
\end{abstract}

%% file: chapters/01_intro.tex
\section{Introduction}
Our goal is to accurately estimate the 3D pose and shape of a person from monocular video.
This is a longstanding problem and, while the field has made rapid progress, several key challenges remain.
First, human motion should be computed in a consistent global coordinate system.
Second, the method should be computationally efficient. Third, the results should be accurate, temporally smooth, detailed, natural looking, and have realistic foot-ground contact. Fourth, the capture should work with an arbitrary moving camera. These constraints need to be satisfied to make markerless human motion capture widely available for applications in gaming, AR/VR, autonomous driving, sports analysis, and human-robot interaction. We address these challenges with {\em WHAM (World-grounded Humans with Accurate Motion)}, which enables fast and accurate recovery of 3D human motion from a moving camera; see Fig.~\ref{fig:teaser}.

It seems natural that, in estimating 3D humans from video, we should be able to exploit the temporal nature of video.
Counter-intuitively, existing video-based methods for 3D human pose and shape (HPS) estimation \cite{hmmr, vibe, meva, tcmr, mpsnet, glot} are less accurate than the best single-frame methods \cite{hmr, spin, pare, pymafx, hybrik, pymafx, cliff, osx, hmr2.0}. This may be an issue of training data. There are large datasets of single images with ground-truth 3D human poses containing a diversity of body shapes, poses, backgrounds, lighting, etc. In contrast, video datasets with ground truth are much more limited.

To address this, WHAM  
leverages both the large-scale AMASS motion capture (mocap) dataset \cite{amass} and video datasets. Our key idea is to learn about 3D human motion from AMASS and then learn to fuse this information with temporal image cues from video, getting the best of both. 
Similar to previous work \cite{motionbert, proxycap}, we use AMASS to generate synthetic 2D keypoints and ground-truth motion sequence pairs, from which we pretrain a motion encoder, which captures the {\em motion context}, and decoder that {\em lifts}  sequences of 2D keypoints to sequences of 3D poses. 
Given the robustness of recent 2D keypoint detection models \cite{vitpose, darkpose}, our pretrained model does a good job of predicting human pose from video.

Keypoints alone, however, are too sparse for accurate 3D mesh estimation. To improve accuracy, we jointly train a feature integrator network that merges information from video and 2D-keypoint sequences. We use a pretrained image encoder from previous work \cite{spin, hmr2.0, cliff, bedlam} and train the feature integrator using video datasets \cite{3dpw, human36m, hmmr, mpii3d}.
This integration process supplements the motion context extracted from the sparse 2D keypoints with dense visual context, significantly improving the recovered pose and shape accuracy.

While the above approach produces accurate motion, we want this motion in global coordinates, unlike most previous methods that compute the body in camera coordinates. Estimating the global human trajectory is challenging when the camera is moving because the motions of the body and the camera are entangled. Recent work addresses this with optimization fitting based on a learned human motion prior and camera information from SLAM methods \cite{slahmr, smartmocap, pace} or dense 3D scene information from COLMAP \cite{4dcapture}. However, these methods are computationally expensive and far from real time. Recent regression-based methods are faster but either constrain the problem with static or known camera conditions \cite{proxycap, glopro} or have temporal jitter and limited accuracy \cite{trace}. We tackle this challenge with two additional modules. First, we predict the global orientation and root velocity of the human from the sequence of 2D keypoints by training a {\em global trajectory decoder}. Specifically, we concatenate the camera's angular velocity to the context and train the global trajectory decoder to recursively predict the current orientation and root velocity, effectively factoring camera motion from human motion.
WHAM takes the camera's angular velocity either from the output of a SLAM method or from a camera's gyroscope when available. 

The above solution relies on knowledge of human motion learned from AMASS.
Therefore, it can fail to capture elevation changes when the surface is not flat, e.g.~when ascending the stairs because AMASS has a limited amount of such data.
To address this, we introduce foot contact as an additional explicit form of motion context. We train WHAM to predict the likelihood of foot-ground contact using estimated contact labels from both AMASS and 3D video datasets. We then train a trajectory refinement network that outputs an update to the root orientation and velocity based on the information about the foot contact/velocity. This refinement enables WHAM to accurately estimate human motion in a global coordinate system even when the terrain is not flat.

WHAM has very low computational overhead because it is an on-line algorithm that recursively predicts the pose, shape, and global motion parameters. The network, excluding preprocessing (bounding box detection, keypoint detection, and person identification), runs at 200 fps, significantly faster than prior methods. Also, despite not using global optimization like \cite{slahmr}, we obtain accurate 3D camera trajectories and global body motions with minimal drift. Through extensive comparisons on multiple in-the-wild datasets as well as detailed ablation studies, we find that WHAM achieves state-of-the-art (SOTA) accuracy on 3D human pose estimation as well as global trajectory estimation (see Fig.~\ref{fig:teaser}).

In summary, in this paper we: (1) introduce the first approach to effectively fuse 3D human motion context and video context for 3D HPS regression;  
(2)  propose a novel global trajectory estimation framework that leverages motion context and foot contact to effectively address foot sliding and enable the 3D tracking of people on non-planar surfaces; 
(3) efficiently perform HPS regression in global coordinates;
(4)  achieve state-of-the-art (SOTA) performance on multiple in-the-wild benchmark datasets (3DPW \cite{3dpw}, RICH \cite{rich}, EMDB \cite{emdb}). 
WHAM is the first video-based method to outperform all image-based and video-based methods on per-frame accuracy 
while maintaining temporal smoothness. 

%% file: chapters/02_related.tex
\section{Related Work}
\noindent\textbf{Image-based 3D HPS Estimation.} 
There are two broad classes of methods for recovering 3D HPS from images: model-free \cite{i2lmeshnet, graphcmr, mpt} and model-based \cite{hmr,spin,hmr2.0,hmr-eft, eft, neuralannot, cliff, 3Dpseudpgts}.
Here we focus on model-based methods, which estimate the low-dimensional parameters of a statistical body model \cite{smpl, smplx, star, ghum}. 
While early work explores optimization-based methods \cite{smplify}, here we focus on direct regression methods based on deep learning.

Many existing methods follow the architecture of HMR \cite{hmr}, which uses a pretrained backbone to predict image features followed by a multilayer perceptron (MLP) that regresses SMPL \cite{smpl} pose parameters from image features.
Training such networks typically leverages paired images with SMPL parameters; these parameters are often pseudo-groundtruth (p-GT),  estimated from 2D keypoints \cite{spin, eft, neuralannot, 3Dpseudpgts, cliff}.
Other architectures for HPS regression have also been proposed
\cite{pare, hybrik, pymafx, niki, hmr2.0, spec}.
None of these methods use video or estimate the body in global coordinates. While quite accurate, when these image-based models are 
applied independently to frames of a video sequence, the shape and pose can be temporally inconsistent. 
In contrast, WHAM effectively aggregates temporal information to provide frame-accurate and temporally-coherent 3D HPS estimation.

\noindent\textbf{Video-Based 3D HPS.} 
Video-based methods typically encode temporal information by combining static features extracted by a backbone from each frame.  
HMMR \cite{hmmr} uses a convolutional encoder, while VIBE \cite{vibe} and MEVA \cite{meva} employ recurrent neural networks. 
TCMR \cite{tcmr} divides sequences into past, future, and whole frames, aggregating information to strongly constrain the output with motion consistency. MPS-Net \cite{mpsnet} uses attention to capture non-local motion context and a hierarchical architecture to aggregate temporal features. Both MAED \cite{maed} and GLoT \cite{glot} use transformer architectures \cite{transformer} to encode videos. MAED encodes videos in both temporal (across frames) and spatial (within each frame) dimensions and leverages the kinematic tree to iteratively regress each joint angle.
GLoT encodes long-term temporal correlations 
and refines local details by 
focusing on nearby frames. 
Despite integrating information across frames, all existing video-based methods have lower accuracy than the best single-frame methods. 

Given limited video training data with ground truth SMPL poses, several single-frame methods infer a mesh from 2D/3D keypoints \cite{pose2mesh, holopose, pavlakos, virtualmarker, i2lmeshnet} and use the keypoints as a proxy for training.
Another set of approaches exploits 3D mocap data, which is plentiful \cite{amass}, to train a network to lift 2D joints to 3D poses, which are used as a proxy for ground truth 3D.
MotionBERT \cite{motionbert} synthesizes 2D keypoints through orthographic projection to learn a unified motion representation. 
ProxyCap \cite{proxycap} projects synthetic 3D keypoints into virtual cameras using a heuristic camera pose distribution. 
Despite benefiting from the scale of mocap datasets, these approaches do not fully utilize the visual information available in the video at test time. Here, we propose a combined network architecture and training strategy that leverages both proxy representations of human pose (lifting) and visual context extracted from video. 

\noindent\textbf{Global 3D Motion Estimation with Dense Sensors.} 
Several methods augment video data with other sensors to estimate 3D HPS in world coordinates.
The 3DPW dataset \cite{3dpw} employs pre-calibrated body-worn inertial sensors and a handheld camera to jointly optimize the camera and human motion in challenging environments. 
Similarly, the EMDB dataset \cite{emdb} uses electromagnetic sensors with an RGB-D camera, enabling accurate human motion capture in the world. 
While body-worn sensors aid global human motion reconstruction, they are intrusive, require cooperation, and do not help with archival video.
BodySLAM++ \cite{bodyslam++} uses an optimization method with a 
visual-inertial sensor, comprising stereo cameras and an IMU. 
In contrast, we use a standard monocular camera,  balancing accessibility and accuracy without using specialized equipment.
While WHAM can take the camera gyro as input, this is not required.

\noindent\textbf{Monocular Global 3D Human Trajectory Estimation.} 
Estimating the global human trajectory from a monocular dynamic camera is challenging.
Previous work relies on learned prior distributions of human motion to separate human motion from camera motion. GLAMR \cite{glamr} computes the global trajectory based on a predicted and infilled 3D motion sequence and optimizes it across multiple individuals in the scene. However, since GLAMR does not consider camera motion cues, the output trajectory may be noisy when the camera is rotating. SLAHMR \cite{slahmr} and PACE \cite{pace} use off-the-shelf SLAM algorithms \cite{droid, dpvo} and jointly optimize the camera and human motion to minimize the negative log-likelihood of a learned motion prior \cite{humor}. 
While they achieve good results, their optimization approach is computationally expensive. TRACE \cite{trace} is a pure regression method that utilizes optical flow as a motion cue and estimates multiple people at once, but
lacks temporal consistency. GloPro \cite{glopro} regresses the uncertainty 
of the global human motion in real-time, but requires known camera poses.
In contrast, WHAM leverages both explicit and implicit prior knowledge of human motion and efficiently reconstructs accurate and temporally coherent 3D human motion in world coordinates.

%% file: chapters/03_method.tex
\input{figures/02_Architecture}

\section{Methods}
\subsection{Overview}
An overview of our World-grounded Human with Accurate Motion (WHAM) framework is illustrated in Fig.~\ref{fig:overview}. The input to WHAM is a raw video data $\{I^{(t)}\}_{t=0}^{T}$, captured by a camera with possibly unknown motion. Our goal is to predict the corresponding sequence of the SMPL model parameters $\{ \Theta^{(t)} \}_{t=0}^{T}$, as well as the root orientation $\{\Gamma^{(t)}\}_{t=0}^{T}$ and translation $\{\tau^{(t)}\}_{t=0}^{T}$, expressed in the world coordinate system. We use ViTPose \cite{vitpose} to detect 2D keypoints $\{x^{(t)}_{2D}\}_{t=0}^{T}$ from which we obtain motion features $\{\phi_m^{(t)}\}_{t=0}^{T}$ using the motion encoder. Additionally, we use a pretrained image encoder \cite{spin, cliff, hmr2.0} to extract static image features $\{\phi_i^{(t)}\}_{t=0}^{T}$ and integrate them with $\{\phi_m^{(t)}\}_{t=0}^{T}$ to obtain fine-grained motion features $\{\hat{\phi}_m^{(t)}\}_{t=0}^{T}$ from which we regress 3D human motion in the world coordinate system.

\subsection{Network Architecture}
\noindent\textbf{Uni-directional Motion Encoder and Decoder.} In contrast to existing methods \cite{tcmr, meva, mpsnet, glot, motionbert}, which use windows with a fixed time duration, we use uni-directional recurrent neural networks (RNN) for the motion encoder and decoder, making WHAM suitable for online inference. The objective of the motion encoder, $E_M$, is to extract the motion context, $\phi_m^{(t)}$, from the current and previous sequence of 2D keypoints and the initial hidden state, $h_E^{(0)}$:
\[
\phi_m^{(t)} = E_M\big{(}x_{2D}^{(0)}, x_{2D}^{(1)}, ..., x_{2D}^{(t)} | h_{E}^{(0)} \big{)}.
\]
We normalize keypoints to a bounding box around the person and concatenate the box's center and scale to the keypoints, similar to CLIFF \cite{cliff}. 
The role of the motion decoder, $D_M$, is to recover SMPL parameters, $(\theta, \beta)$, weak-perspective camera translation, $c$, and foot-ground contact probability, $p$, from the motion feature history:
\[
\big{(}\theta^{(t)}, \beta^{(t)}, c^{(t)}, p^{(t)} \big{)}= D_M\big{(} \hat{\phi}_m^{(0)}, ..., \hat{\phi}_m^{(t)} | h^{(0)}_{D} \big{)}.
\]
\noindent Here, $\hat{\phi}_m^{(t)}$ is the motion feature integrated with the image feature $\phi_i^{(t)}$ (described below). 
During pretraining on synthetic data, the image feature is not available and we set $\hat{\phi}_m^{(t)} = \phi_m^{(t)}$. 
As the encoder and decoder are tasked with reconstructing a dense 3D representation $\Theta$ from a sparse 2D input signal $x_{2D}$, we design an intermediate task to predict the 3D keypoints $x_{3D}$ and use them as the intermediate motion representation. This cascaded approach guides $\phi_m$ to represent the implicit context of motion and the 3D spatial structure of the body.
Similar to PIP \cite{pip}, we use Neural Initialization that uses MLP to initialize the hidden state of the motion encoder and decoder, $\big{(}h_E^{(0)}, h_D^{(0)}\big{)}$; see section \ref{ablation} and Sup.~Mat.~for details.

\noindent\textbf{Motion and Visual Feature Integrator.} 
We use the AMASS dataset to synthetically generate 2D sequences by projecting 3D SMPL joints into images with varied camera motions.
This provides effectively limitless training data that is
far more diverse than existing video datasets that contain ground truth 3D pose and shape.
Although we leverage the temporal human motion context, lifting 2D keypoints to 3D meshes is an ambiguous task.
A key idea is to augment this 2D keypoint information with image cues that can help disambiguate the 3D pose.  
Specifically, we use an image encoder \cite{spin, cliff, bedlam, hmr2.0}, pretrained on the human mesh recovery task, to extract image features $\phi_i$, which contain dense visual contextual information related to the 3D human pose and shape.
We then train a feature integrator network, $F_I$, to combine $\phi_m$ with $\phi_i$, integrating motion and visual context.
The feature integrator uses a simple yet effective residual connection:
\[
\hat{\phi}_m^{(t)} = \phi_m^{(t)} + F_I\Big{(}\text{concat} \big{(} \phi_m^{(t)}, \phi_i^{(t)} \big{)} \Big{)}.
\]
This supplements motion features pretrained on the 2D-to-3D lifting task using AMASS with visual context, resulting in enriched motion features that use image evidence to help disambiguate the task.

\noindent\textbf{Global Trajectory Decoder.} We design an additional decoder, $D_T$, to predict the rough global root orientation $\Gamma_0^{(t)}$ and root velocity $v_0^{(t)}$ from the motion feature $\phi_m^{(t)}$. Since $\phi_m$ is derived from the input signals in the camera coordinates, it is highly challenging to decouple the human and camera motion from it. To address this ambiguity, we append the angular velocity of the camera, $\omega^{(t)}$, to the motion feature, $\phi_m^{(t)}$, to create a camera-agnostic motion context. This design choice makes WHAM compatible with both off-the-shelf SLAM algorithms \cite{dpvo, droid} and gyroscope measurements that are widely available from modern digital cameras. 
We recursively predict global orientation, $\Gamma_0^{(t)}$, using the uni-directional RNN.
\[
\big{(} \Gamma_0^{(t)}, v_0^{(t)} \big{)} = D_T \big{(} \phi_m^{(0)}, \omega^{(0)}, ..., \phi_m^{(t)},\omega^{(t)}\big{)}.
\]

\noindent\textbf{Contact Aware Trajectory Refinement.} Good 3D motion in world coordinates in most scenarios implies accurate foot-ground contact without sliding.
We want WHAM to generalize beyond flat ground planes, which are typically assumed in prior work.
Specifically, our new trajectory refiner aims to resolve foot sliding and enables WHAM to generalize well to diverse motions, including climbing stairs. The refinement involves two stages. First, we adjust the ego-centric root velocity to $\tilde{v}^{(t)}$ to minimize foot sliding, based on the foot-ground contact probability $p^{(t)}$, estimated from the motion decoder $D_M$:
\[
\tilde{v}^{(t)} = v_0^{(t)} - \big{(} \Gamma_0^{(t)} \big{)}^{-1} \bar{v}_f^{(t)},
\]
where $\bar{v}_f^{(t)}$ is the averaged velocity of the toes and heels in world coordinates when their contact probability, $p^{(t)}$, is higher than a threshold. However, this velocity adjustment often introduces noisy translation when the contact and pose estimation is inaccurate. Therefore, we propose a simple learning mechanism in which a trajectory refining network, $R_T$, updates the root orientation and velocity to address this issue. Finally, the global translation is computed through a roll-out operation:
\begin{eqnarray*}
\big{(} \Gamma^{(t)}, v^{(t)} \big{)} & = & R_T\big{(} \phi_m^{(0)}, \Gamma_0^{(0)}, \Tilde{v}^{(0)}, ..., \phi_m^{(t)}, \Gamma_0^{(t)}, \Tilde{v}^{(t)} \big{)},\\
\tau^{(t)} & = & \sum_{i=0}^{t-1}{\Gamma^{(i)} v^{(i)}}.
\end{eqnarray*}

In summary, this full process reconstructs accurate 3D human pose and shape in both the camera and world coordinates from a single monocular video sequence (Fig. \ref{fig:overview}).

\input{figures/03_TrainingScheme}
\subsection{Training}
\noindent\textbf{Pretraining on AMASS.}  We train in two stages:  (1) pretraining with synthetic data, and (2) fine-tuning with real data (Fig. \ref{fig:trainingscheme}). The objective of the pretraining stage is to teach the motion encoder to extract motion context from the input 2D keypoint sequence. The motion and trajectory decoders then map this motion context to the corresponding 3D motion and global trajectory spaces (i.e.~they lift the encoding to 3D). 
We use the AMASS dataset \cite{amass} to generate an extensive set of synthetic pairs consisting of sequences of 2D keypoints together with the ground truth SMPL parameters. 

To synthesize 2D keypoints from AMASS, we create virtual cameras onto which we project 3D keypoints derived from the ground truth mesh. Unlike MotionBERT \cite{motionbert} and ProxyCap \cite{proxycap}, which use static cameras for keypoint projection, we employ dynamic cameras that incorporate both rotational and translational motion. 
This choice has two main motivations.
First, it accounts for the inherent differences between human motions captured in static and dynamic camera setups. Second, it enables the learning of a camera-agnostic motion representation, from which the trajectory decoder can reconstruct the global trajectory. 
We also augment the 2D data with noise and masking.
For details of the synthetic generation process see Sup.~Mat.

\noindent\textbf{Fine-tuning on Video Datasets.} 
Starting with the pretrained network, 
we fine-tune WHAM on four video datasets: 3DPW \cite{3dpw}, Human3.6M \cite{human36m},  MPI-INF-3DHP \cite{mpii3d}, and InstaVariety \cite{hmmr}. For the human mesh recovery task, we supervise WHAM on ground-truth SMPL parameters from AMASS and 3DPW, 3D keypoints from Human3.6M and MPI-INF-3DHP, and 2D keypoints from InstaVariety. For the global trajectory estimation task, we use AMASS, Human3.6M, and MPI-INF-3DHP. Additionally, during training we experiment with adding BEDLAM \cite{bedlam}, a large synthetic dataset with realistic video and ground truth SMPL parameters.

The fine-tuning has two objectives: 1) exposing the network to real 2D keypoints, instead of training it solely on synthetic data, and 2) training the feature integrator network to aggregate motion and image features. To achieve these goals, we jointly train the entire network on the video datasets while setting a smaller learning rate on the pretrained modules (see Fig.~\ref{fig:trainingscheme}). Consistent with prior work \cite{vibe, meva, tcmr, mpsnet, glot}, we employ a pretrained and fixed-weight image encoder \cite{spin} to extract image features. However, to leverage recent network architectures and training strategies, we also experiment with different types of encoders \cite{cliff, bedlam, hmr2.0} 
in the following section.

\noindent\textbf{Implementation Details.} For the pretraining stage, we train WHAM on AMASS for 80 epochs with the learning rate of $5\times 10^{-4}$. Then we finetune WHAM on 3DPW, MPI-INF-3DHP, Human3.6M, and InstaVariety for 30 epochs. We use a learning rate of $1\times 10^{-4}$ for the feature integrator and $1\times 10^{-5}$ for the pretrained components during finetuning. During training, we use the Adam optimizer and a batch size of 64 and split sequences into 81-frame chunks.

%% file: figures/02_Architecture.tex
\begin{figure*}[ht!]
\centerline{\includegraphics[width=0.9\linewidth]{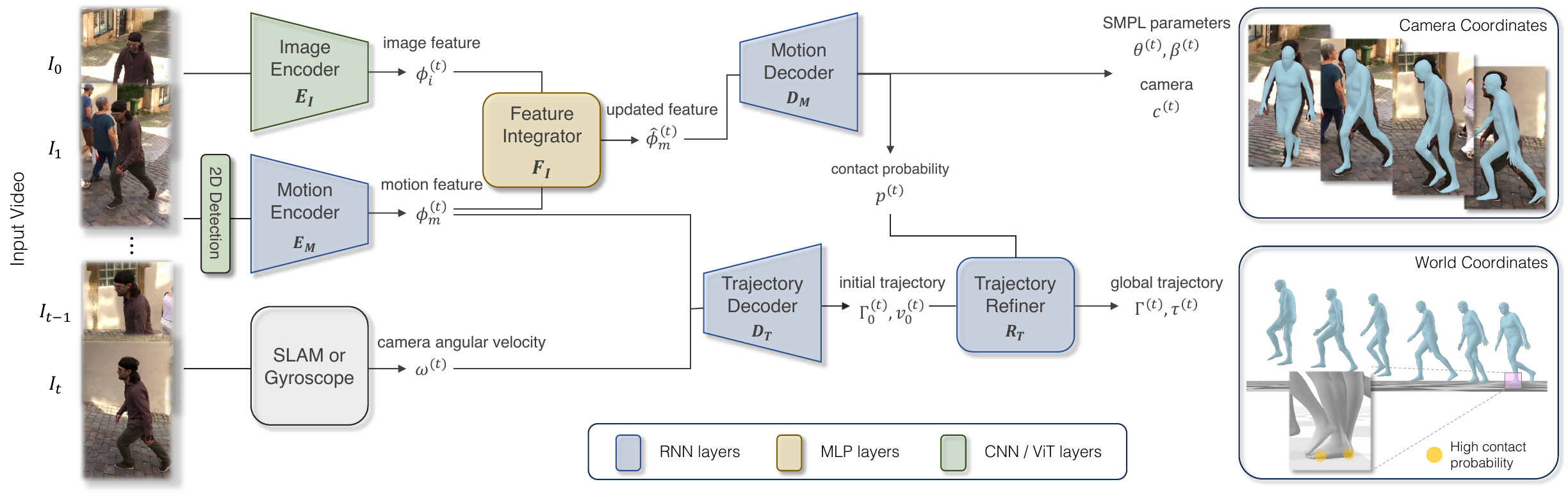}}
\vspace{-0.1in}
   \caption{\textbf{An Overview of WHAM.} WHAM takes the sequence of 2D keypoints estimated by a pretrained detector and encodes it into a motion feature. WHAM then updates the motion feature using another sequence of image features extracted from the image encoder through the feature integrator. From the updated motion feature, the Local Motion Decoder estimates 3D motion in the camera coordinate system and foot-ground contact probability. The Trajectory Decoder takes the motion feature and camera angular velocity to initially estimate the global root orientation and egocentric velocity, which are then updated through the Trajectory Refiner using the foot-ground contact. The final output of WHAM is pixel-aligned 3D human motion with the 3D trajectory in the global coordinates.}
\vspace*{-0.2cm}
\label{fig:overview}
\end{figure*}


%% file: figures/03_TrainingScheme.tex
\newcommand*{\inlineimg}[1]{%
    \raisebox{-.1\baselineskip}{%
        \includegraphics[
        height=0.7\baselineskip,
        width=0.7\baselineskip,
        keepaspectratio,
        ]{#1}%
    }%
}

\begin{figure}[t]
\centerline{\includegraphics[width=1.0\linewidth]{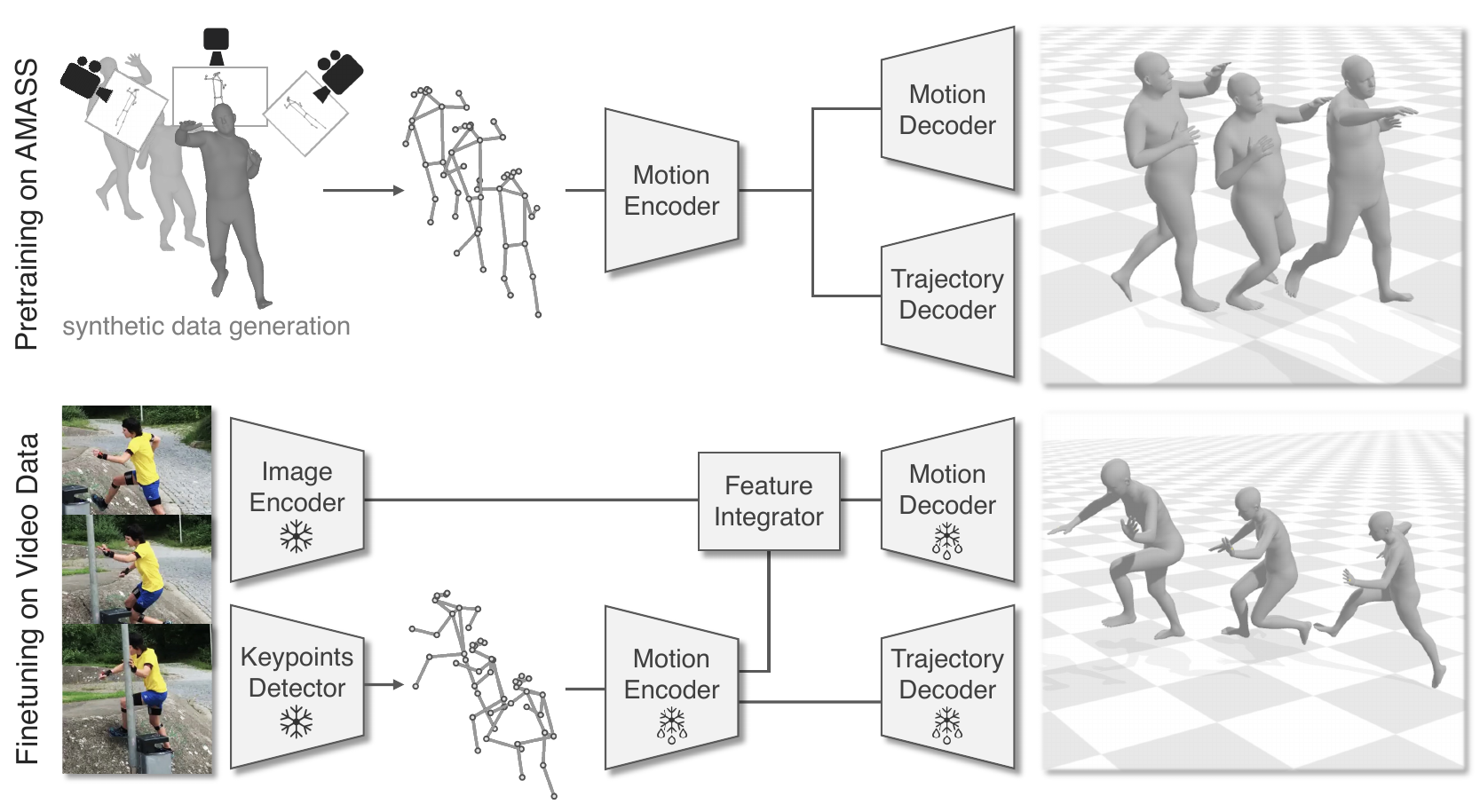}}
\vspace{-0.1in}
   \caption{\textbf{WHAM's Two-Stage Training Scheme.} During pre-taining, we generate synthetic 2D keypoint sequences from AMASS \cite{amass} and train a motion encoder and decoder on the generated data (\textit{top}). 
   We then leverage video datasets with ground truth SMPL parameters, for which there is much less data. We use the fixed-weight pre-trained image encoder and keypoints detector (\inlineimg{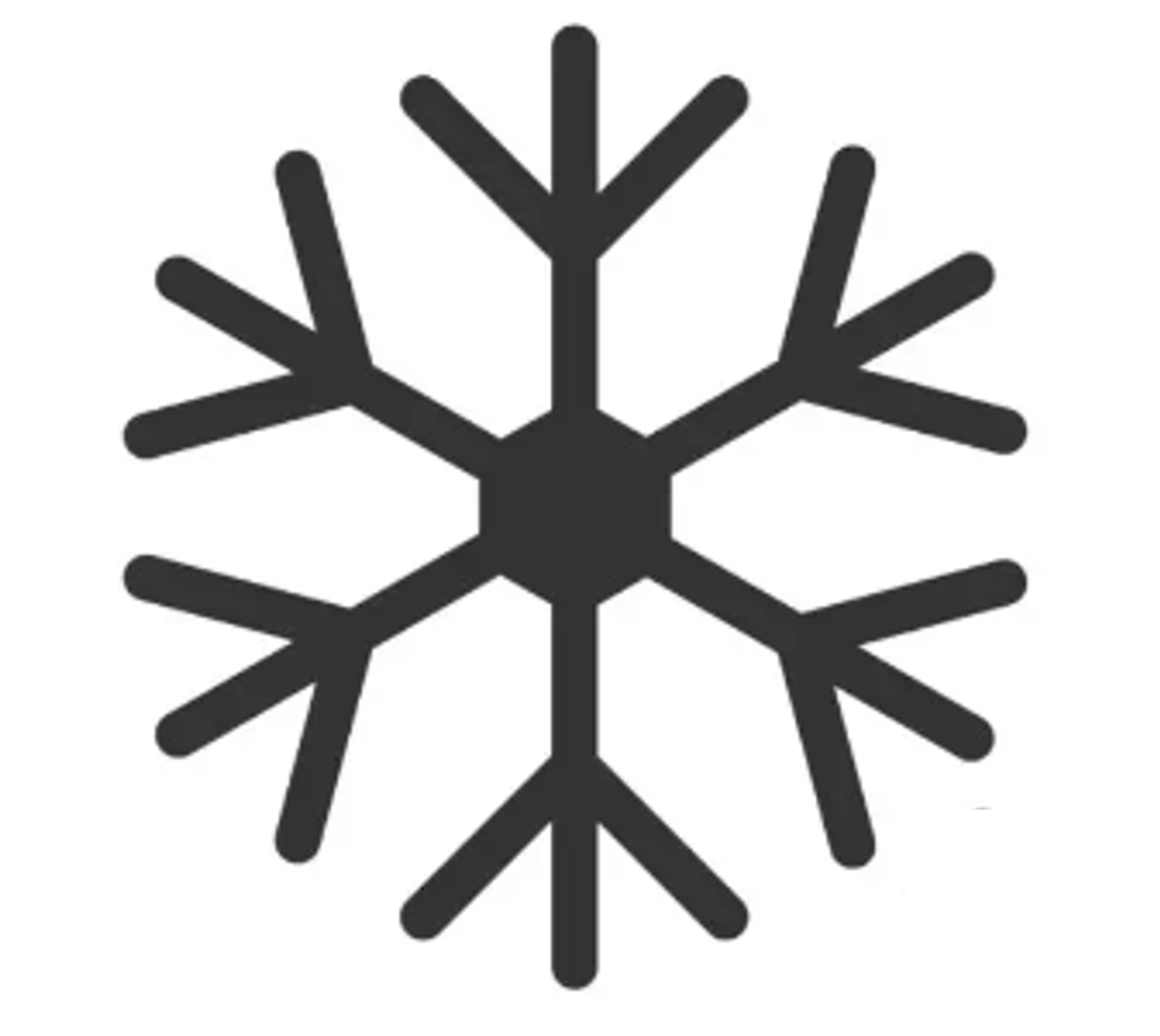}) to extract image features and 2D keypoints.
   In this stage, we train the feature integration network while fine-tuning the motion encoder and motion/trajectory decoders, marked \inlineimg{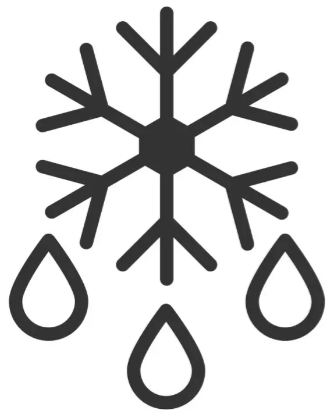} (\textit{bottom}).}
\vspace*{-0.2cm}
\label{fig:trainingscheme}
\end{figure}


%% file: chapters/04_experiments.tex
\section{Experiments}


\noindent\textbf{Datasets.}
We evaluate WHAM on three in-the-wild benchmarks: 3DPW \cite{3dpw}, RICH \cite{rich}, and EMDB \cite{emdb}. Following previous work \cite{hmr, spin, vibe, tcmr, meva, cliff, bedlam}, we perform the evaluation in camera coordinates. The estimated global trajectory is evaluated on a subset of EMDB (\textit{EMDB 2}) for which they provide ground truth global motion with dynamic cameras (used for evaluation). 
We also test on new sequences captured using an iPhone with the gyroscope. See Sup.~Mat.~for more details on the datasets and iPhone results.

\noindent\textbf{Evaluation metrics.} To evaluate the accuracy of 3D human pose and shape estimation, we compute Mean Per Joint Position Error (MPJPE), Procrustes-aligned MPJPE (PA-MPJPE), and Per Vertex Error (PVE) measured in millimeters $(mm)$. We compute Acceleration error (Accel, in $m/s^2$)\footnote{Previous work follows \cite{hmmr} in reporting Accel in $mm/\mathrm{frame}^2$. To remove the dependency on frame rate, we convert all previous results to  $m/s^2$.} 
to measure the inter-frame smoothness of the reconstructed motion. We also evaluate the motion reconstruction and trajectory estimation accuracy in the world-frame. 
Following previous work \cite{slahmr, pace}, we split sequences into smaller segments of 100 frames and align each output segment with the ground-truth data using the first two frames (W-MPJPE$_{100}$) or the entire segment (WA-MPJPE$_{100}$) in $mm$. 
These previous metrics give an unrealistic picture of 3D performance as they do not measure drift over long sequences.
Therefore, we also evaluate the error over the {\em entire trajectory} after the rigid alignment and measure Root Translation Error (RTE in \%) normalized by the actual displacement of the person. We also assess the jitter of the motion in the world coordinate system in $10m/s^3$ and foot sliding during the contact (FS in $mm$).

\input{tables/finals/Table1_1}
\input{tables/finals/Table1_2}

\subsection{3D Human Motion Recovery}
\noindent\textbf{Per-frame accuracy.} Table \ref{tab:table1} presents a comprehensive comparison of WHAM and the existing state-of-the-art per-frame and video-based methods across three benchmark datasets \cite{3dpw, rich, emdb}. 
Because none of the methods are exposed to data from RICH or EMDB during training, results on these datasets are indicative of each method's ability to generalize.
WHAM (Res), WHAM (HR), and WHAM (ViT) correspond to different architectures for the pretrained image encoders, derived from SPIN (ResNet-50) \cite{spin}, CLIFF (HRNet-W48) \cite{cliff, bedlam}, and HMR2.0 (ViT-H/16) \cite{hmr2.0}, respectively. 
Not surprisingly, WHAM (HR) is more accurate than WHAM (Res), while the transformer-based version, WHAM (ViT), is the most accurate. The backbone matters, with WHAM (ViT) outperforming all previous methods on all per-frame metrics (MPJPE, PA-MPJPE, and PVE) on all benchmarks. Even with the simplest ResNet backbone, WHAM (Res) outperforms every method except for ReFit on RICH.

Training on BEDLAM consistently improves accuracy in prior work (see \cite{Cai:NeurIPS:2023,bedlam,refit}), and we find the same here as shown in Table \ref{tab:table1_2}. However, compared to ReFit, WHAM exhibits relatively smaller performance improvement by adding BEDLAM. One possible reason for this is that, unlike ReFit, we did not finetune the image encoder during training to limit training time.


\input{figures/04_QualitativeHPS}

\noindent\textbf{Inter-frame smoothness.} We also evaluate the inter-frame smoothness using the acceleration error. Compared with state-of-the-art per-frame methods \cite{hmr2.0, cliff, bedlam, hybrik}, WHAM has significantly lower acceleration error. This indicates that WHAM reconstructs smooth and more plausible 3D human motion across frames while not sacrificing high per-frame accuracy. On the other hand, when compared to recent temporal methods \cite{glot, mpsnet, tcmr}, WHAM exhibits comparable or slightly higher acceleration error. However, we observe that these video-based methods tend to over-smooth the human motion, resulting in lower accuracy on per-frame metrics. 

To provide intuition for these numbers, we qualitatively compare WHAM with TCMR \cite{tcmr} and GLoT \cite{glot} in Fig.~\ref{fig:qualitativeHPS}. While producing smooth results, TCMR and GLoT fail to capture the bending of the left knee when the subject is ascending the stairs, while WHAM more accurately reconstructs the 3D human pose.

\input{tables/finals/Table2}

\subsection{3D Global Trajectory Recovery}
To evaluate global trajectory recovery, we compare WHAM with the state-of-the-art methods and a baseline that combines a SLAM method  (DPVO \cite{dpvo}) and a per-frame method (HMR2.0 \cite{hmr2.0}); see Table \ref{tab:table2}. WHAM is agnostic to the source of the camera angular velocity and we compare results using DPVO, DROID-SLAM \cite{droid} and the ground truth angular velocity (gyro).

As shown in Table \ref{tab:table2}, WHAM outperforms the existing methods on all metrics. Specifically, combining WHAM with DPVO is more accurate than the global trajectory estimation of DPVO combined with HMR2.0, illustrating that our method actively refines the global trajectory instead of performing a simple integration. 
DROID-SLAM gives slightly better results than DPVO.
Furthermore, WHAM significantly outperforms the regression-based method, TRACE, on jitter and foot sliding metrics. We further demonstrate this in Figs.~\ref{fig:qualitativeGlobal} and \ref{fig:teaser}, where WHAM captures more consistent and plausible human motion in the global coordinate system than TRACE and SLAHMR for videos captured by dynamic cameras. As depicted in Fig.~\ref{fig:qualitativeTraj}, WHAM outperforms GLAMR, TRACE, and SLAHMR in capturing the pattern of human motion in the global coordinate system.

\input{figures/05_QualitativeGlobal}
\input{figures/06_QualitativeTraj}

\subsection{Ablation Study}\label{ablation}
To provide further insight into our approach, we conduct ablation studies to analyze the contribution of each component to the performance. As shown in Table \ref{tab:table3}, our entire system (WHAM) outperforms the different variants of WHAM that ablate a single component. To be specific, we first observe that adding feature integration improves both motion and global trajectory estimation accuracy when compared with an ablated version without feature integration (w/o $F_I$). 
Similarly, the removal of the pretraining on the 2D-to-3D lifting task using AMASS \cite{amass} (w/o lifting) shows significant performance degradation. 
WHAM also outperforms the ablation of the Neural Initialization (w/o NI), particularly in the MPJPE metric.
In addition, we experiment with WHAM to decode trajectory solely based on the motion context without using the estimated camera angular velocity (w/o $\omega$). Although this version shows similar performance in predicting 3D human pose, it suffers from the entanglement of camera and human motion, resulting in significantly high overall trajectory error (RTE).
Last, we observe that WHAM without the trajectory refinement (w/o traj.~ref.) gives larger global trajectory and foot sliding errors in return for less jitter, indicating that our approach contributes to the global trajectory accuracy and helps reduce foot sliding. Core details are presented here; see Sup.~Mat.~for more details and information on run-time cost.


\input{tables/finals/Table3}

%% file: tables/finals/Table1_1.tex
\newcolumntype{?}{!{\vrule width 0.75pt}}

\begin{table*}[tbh]
\centering
\setlength{\tabcolsep}{3pt}
\renewcommand{\arraystretch}{1.2}
\resizebox{0.8\textwidth}{!}
{\small{
\begin{tabular}{cl?cccc?cccc?cccc}
\cmidrule[0.75pt]{1-14}
& & \multicolumn{4}{c}{3DPW (14)} & \multicolumn{4}{c}{RICH (24)} & \multicolumn{4}{c}{EMDB (24)} \\
\cmidrule(lr){3-6} \cmidrule(lr){7-10} \cmidrule(lr){11-14}

& Models & \scriptsize{PA-MPJPE} & \scriptsize{MPJPE} & \scriptsize{PVE} & \scriptsize{Accel} & \scriptsize{PA-MPJPE} & \scriptsize{MPJPE} & \scriptsize{PVE} & \scriptsize{Accel} & \scriptsize{PA-MPJPE} & \scriptsize{MPJPE} & \scriptsize{PVE} & \scriptsize{Accel} \\
\cmidrule{1-14}

\multirow{6}{1em}{\rotatebox[origin=c]{90}{per-frame}} 
& SPIN \cite{spin} & 59.2 & 96.9 & 112.8 & 31.4 & 69.7 & 122.9 & 144.2 & 35.2 & 87.1 & 140.7 & 166.1 & 41.3 \\

& PARE$^*$ \cite{pare} & 46.5 & 74.5 & 88.6 & -- & 60.7 & 109.2 & 123.5 & -- & 72.2 & 113.9 & 133.2 & -- \\

& CLIFF$^*$ \cite{cliff} & 43.0 & 69.0 & 81.2 & 22.5 & 56.6 & 102.6 & 115.0 & 22.4 & 68.3 & 103.5 & 123.7 & 24.5 \\

& HybrIK$^*$ \cite{hybrik} & 41.8 & 71.6 &  82.3 & -- & 56.4 & 96.8 & 110.4 & -- & 65.6 & 103.0 & 122.2 & -- \\

& HMR2.0 \cite{hmr2.0} & 44.4 & 69.8 & 82.2 & 18.1 & 48.1 & 96.0 & 110.9 & 18.8 & 60.7 & 98.3 & 120.8 & 19.9 \\


& ReFit$^*$ \cite{refit} & 40.5 & 65.3 & 75.1 & 18.5 & 47.9 & 80.7 & 92.9 & 17.1 & 58.6 & 88.0 & 104.5 & 20.7 \\

\cmidrule{1-14}

\multirow{11}{1em}{\rotatebox[origin=c]{90}{temporal}} 

& TCMR$^*$ \cite{tcmr} & 52.7 & 86.5 & 101.4 & 6.0 & 65.6 & 119.1 & 137.7 &  5.0 & 79.8 & 127.7 & 150.2 & 5.3 \\

& VIBE$^*$ \cite{vibe} & 51.9 & 82.9 & 98.4 & 18.5 & 68.4 & 120.5 & 140.2 & 21.8 & 81.6 & 126.1 & 149.9 & 26.5 \\

& MPS-Net$^*$ \cite{mpsnet}  & 52.1 & 84.3 & 99.0 & 6.5 & 67.1 & 118.2 & 136.7 & 5.8 & 81.4 & 123.3 & 143.9 & 6.2 \\

& GLoT$^*$ \cite{glot} & 50.6 & 80.7 & 96.4 & \textbf{6.0} & 65.6 & 114.3 & 132.7 & 5.2 & 79.1 & 119.9 & 140.8 & 5.4 \\

& GLAMR \cite{glamr} & 51.1 & -- & -- & 8.0 & 79.9 & -- & -- & 107.7 & 73.8 & 113.8 & 134.9 & 33.0 \\

& TRACE$^*$ \cite{trace} & 50.9 & 79.1 & 95.4 & 28.6 & -- & -- & -- & -- & 71.5 & 110.0 & 129.6 & 25.5 \\

& SLAHMR \cite{slahmr} & 55.9& -- & -- & -- & 52.5 & -- & -- & 9.4 & 69.7 & 93.7 & 111.3 & 7.1 \\

& PACE \cite{pace} & -- & -- & -- & -- & 49.3 & --  & -- & 8.8 & -- & -- & -- & -- \\

\cmidrule{2-14}
& \textbf{WHAM (Res)}$^*$  & 40.2 & 62.7 & 75.1 & 6.3 & 51.8 & 89.5 & 103.2 & \textbf{5.0} & 57.8 & 84.0 & 99.7 & \textbf{5.2} \\
& \textbf{WHAM (HR)}$^*$  & 39.0 & 62.6 & 74.8 & 6.4 & 49.1 & 84.6 & 96.4 & 5.2 & 57.1 & 85.7 & 103.2 & 5.6 \\
&\textbf{WHAM (ViT)}$^*$ & \textbf{35.9} & \textbf{57.8} & \textbf{68.7} & 6.6 & \textbf{44.3} & \textbf{80.0} & \textbf{91.2} & 5.3 & \textbf{50.4} & \textbf{79.7} & \textbf{94.4} & 5.3 \\

\cmidrule[0.75pt]{1-14}
\end{tabular}
}}
    \vspace{-0.075in}
\caption{Quantitative comparison of state-of-the-art models on the 3DPW \cite{3dpw}, RICH \cite{rich}, and EMDB \cite{emdb} datasets. 
Ordering of per-frame and temporal methods is done separately by descending MPJPE on EMDB (except for PACE). For testing on EMDB, we follow the protocol of \textit{EMDB 1} \cite{emdb}. Parenthesis denotes the number of body joints used to compute MPJPE and PA-MPJPE, and $^*$ denotes models trained with the 3DPW training set.
Bold numbers denote the most accurate method in each column. Accel is in $m/s^2$, all other errors are in $mm$.
}
\label{tab:table1}
\end{table*}

%% file: tables/finals/Table1_2.tex

\begin{table}\centering
    \setlength{\tabcolsep}{3pt}
    \resizebox{0.85\columnwidth}{!}{{\small{
    \begin{tabular}{lc?ccccc}
    \cmidrule{1-6}
    & & \multicolumn{4}{c}{3DPW (14)} \\
    \cmidrule(lr){3-6}
    
    Models & Dataset & PA-MPJPE & MPJPE & PVE & Accel \\
    \cmidrule{1-6}

    \multirow{2}{7em}{CLIFF \cite{cliff, bedlam}} & R & 43.6 & 68.8 & 82.1 & 19.2 \\
     & R+B & 43.0 & 66.9 & 78.5 & 31.0 \\
    \cmidrule{1-6}
    
    \multirow{2}{7em}{ReFit \cite{refit}} & R & 40.5 & 65.3 & 75.1 & 18.5 \\
     & R+B & 38.2 & 57.6 & 67.6 & 21.4 \\
    
    \cmidrule{1-6}
    \multirow{2}{7em}{\textbf{WHAM (ViT)}}
    & R & 35.9 & 57.8 & 68.7 & \textbf{6.6} \\
     & R+B & \textbf{35.7} & \textbf{56.9} & \textbf{67.4} & 6.7 \\
    \cmidrule{1-6}
    \end{tabular}
    }}}
    \vspace{-0.1in}
    \caption{Dataset ablation experiments on 3DPW \cite{3dpw}. R denotes the use of real datasets and B denotes BEDLAM.}
    \label{tab:table1_2}
\end{table}

%% file: figures/04_QualitativeHPS.tex
\begin{figure}[tb]
\centerline{\includegraphics[width=0.95\linewidth]{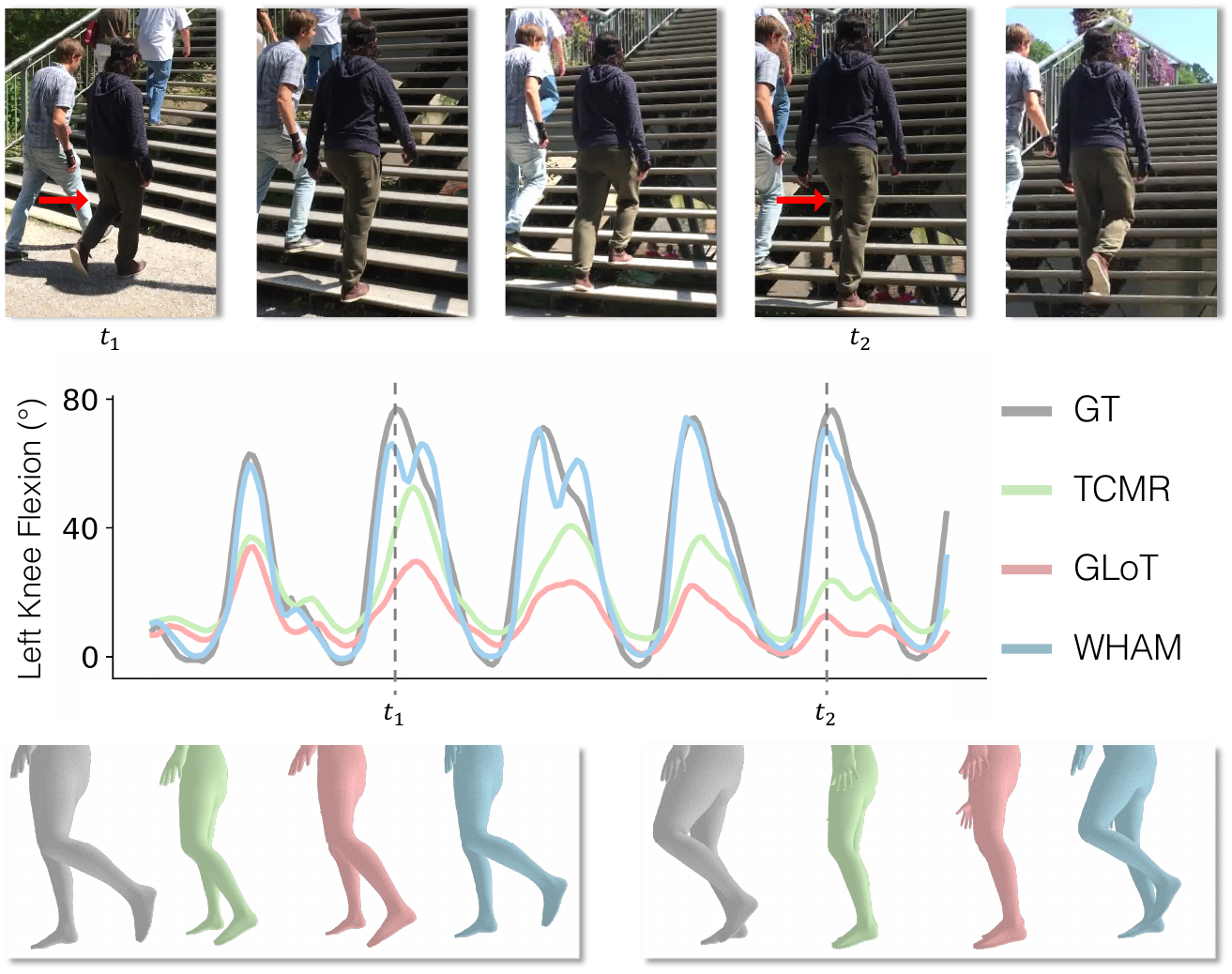}}
\vspace{-0.1in}
   \caption{Qualitative comparison with previous state-of-the-art methods for 3D human pose and shape estimation. See text.}
\vspace*{-0.2cm}
\label{fig:qualitativeHPS}
\end{figure}

%% file: tables/finals/Table2.tex

    \begin{table}\centering
    \setlength{\tabcolsep}{3pt}
    \resizebox{\columnwidth}{!}{{\small{
    \begin{tabular}{l?cccccc}
    \cmidrule{1-7}
    & \multicolumn{6}{c}{EMDB 2} \\
    \cmidrule(lr){2-7}
    
    Models & WA-MPJPE$_{100}$ & W–MPJPE$_{100}$ & RTE & Jitter & FS & \\
    \cmidrule{1-7}
    
    DPVO (+ HMR2.0) \cite{dpvo, hmr2.0} & 647.8 & 2231.4 & 15.8 & 537.3 & 107.6 \\
    GLAMR \cite{glamr} & 280.8 & 726.6 & 11.4 & 46.3 & 20.7 \\
    TRACE \cite{trace} & 529.0 & 1702.3 & 17.7 & 2987.6 & 370.7 \\
    SLAHMR \cite{slahmr} & 326.9 & 776.1 & 10.2 & 31.3 & 14.5 \\
    
    \cmidrule{1-7}
    \textbf{WHAM (w/DPVO \cite{dpvo})} & 135.6 & 354.8 & 6.0 & 22.5 & 4.4 \\
    \textbf{WHAM (w/DROID \cite{droid})} & 133.3 & 343.9 & 4.6 & 21.5 & \textbf{4.4} \\
    \textbf{WHAM (w/ GT gyro)} & \textbf{131.1} & \textbf{335.3} & \textbf{4.1} & \textbf{21.0} & 4.4 \\
    \cmidrule{1-7}
    \end{tabular}
    }}}
        \vspace{-0.1in}
    \caption{Global motion estimation accuracy on EMDB \cite{emdb}.}
    \label{tab:table2}
    \vspace{-0.3em}
    \end{table}

%% file: figures/05_QualitativeGlobal.tex
\begin{figure}[tbh]
\centerline{\includegraphics[width=0.97\linewidth]{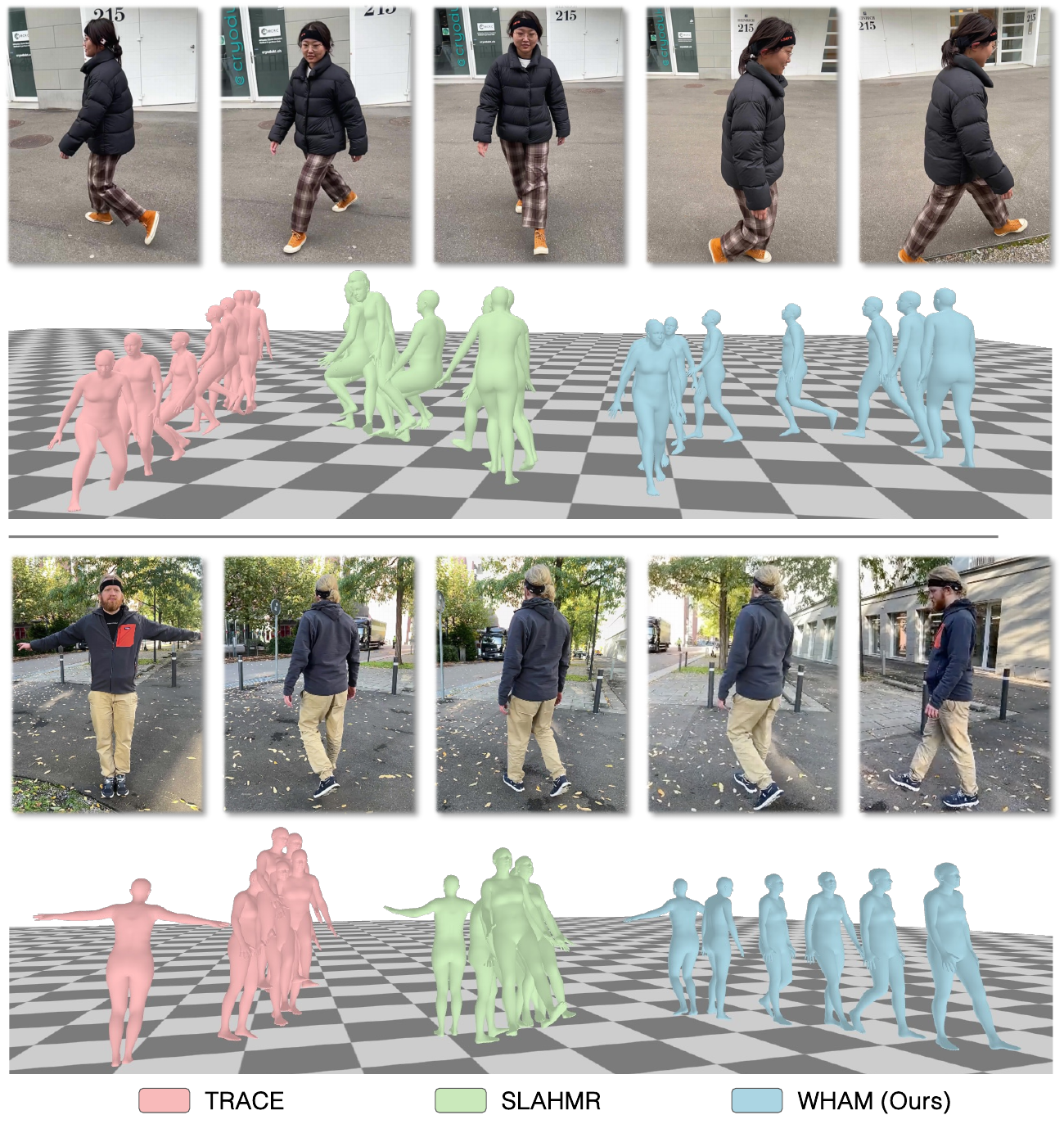}}
\vspace{-0.1in}
   \caption{Qualitative comparison with TRACE \cite{trace} and SLAHMR \cite{slahmr} on global human motion estimation with dynamic cameras.}
\label{fig:qualitativeGlobal}
\end{figure}

%% file: figures/06_QualitativeTraj.tex
\begin{figure}[tbh]
\centerline{\includegraphics[width=1.0\linewidth]{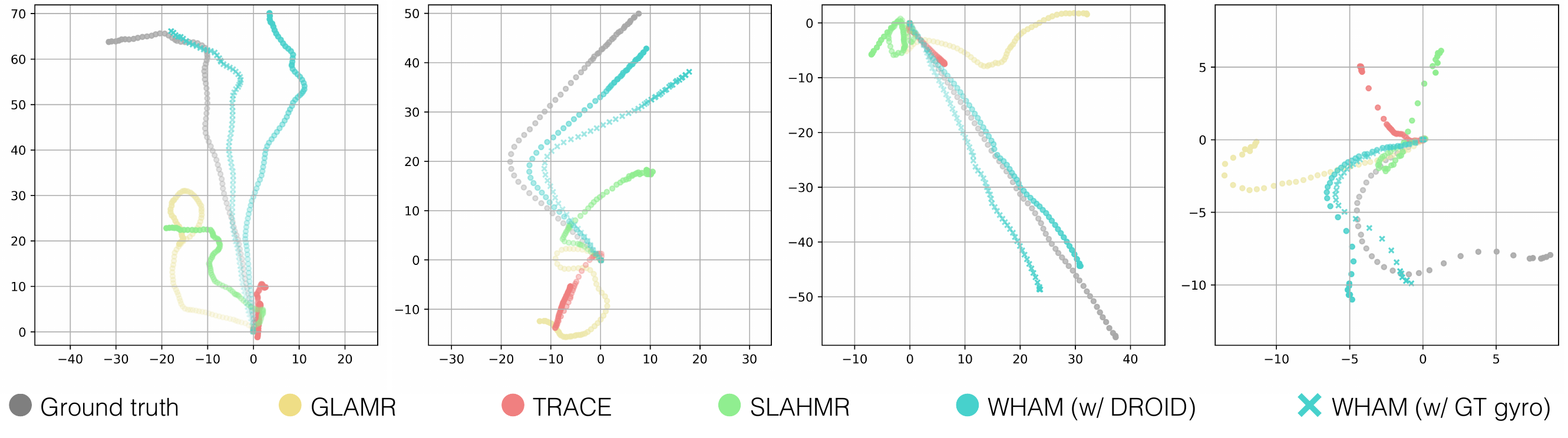}}
\vspace{-0.1in}
   \caption{Comparison of global trajectory estimation on EMDB \cite{emdb}. Overall, WHAM shows better alignment to ground truth data compared to GLAMR \cite{glamr}, TRACE \cite{trace}, and SLAHMR \cite{slahmr}.}
\vspace*{-0.2cm}
\label{fig:qualitativeTraj}
\end{figure}

%% file: tables/finals/Table3.tex

\begin{table}\centering
\setlength{\tabcolsep}{3pt}
\resizebox{\columnwidth}{!}{{\small{
\begin{tabular}{l?ccccccc}
\cmidrule{1-8}
& \multicolumn{7}{c}{EMDB 2} \\
\cmidrule(lr){2-8}

Models & PA–MPJPE & MPJPE & WA–MPJPE$_{100}$ & W–MPJPE$_{100}$ & RTE & Jitter & FS \\
\cmidrule{1-8}

w/o $F_I$ & 44.2 & 69.0 & 147.6 & 377.9 & 6.3 & 23.1 & 5.5 \\
w/o lifting & 60.3 & 83.0 & 238.0 & 693.0 & 11.5 & 24.5 & 5.0 \\
w/o \textit{NI} & 40.4 & 66.3 & 142.7 & 368.1 & 6.8 & 22.3 & 4.6 \\
w/o $\omega$ & 39.1 & 62.0 & 156.5 & 422.0 & 10.1 & 22.1 & 5.0 \\
w/o traj. ref. & \textbf{38.2} & \textbf{59.3} & 154.7 & 407.5 & 6.3 & \textbf{18.8} & 6.5 \\
\textbf{WHAM} (Ours) & \textbf{38.2} & \textbf{59.3} & \textbf{135.6} & \textbf{354.8} & \textbf{6.0} & 22.5 & \textbf{4.4} \\
\cmidrule{1-8}
\end{tabular}
}}}
    \vspace{-0.1in}
\caption{Ablation experiments. See text.}
\label{tab:table3}
\vspace{-0.3em}
\end{table}

%% file: chapters/05_conclusion.tex
\section{Conclusion}
\label{sec:conclusion}
WHAM is a new method to recover accurate 3D human motion in global coordinates from a moving camera more efficiently and accurately than the state-of-the-art approaches.
Our approach leverages the AMASS dataset to train a network to recursively lift 2D sequences of keypoints to sequences of 3D SMPL parameters.
But keypoints alone lack valuable information about the body and its movement.
Consequently, we integrate image context information over time and learn to combine it with the motion context to better estimate human body shape and pose.
Additionally, our method takes an estimate of the camera angular velocity, which can either be computed from a SLAM method or from the camera's gyro when available.
Finally, we combine all this information with an estimate of foot contact to recover the 3D human motion in global coordinates from a monocular video sequence.
WHAM significantly outperforms the existing state-of-the-art methods (both image-based and video-based) on challenging in-the-wild benchmarks in both 3D HPS and the world-coordinate trajectory estimation accuracy.
Because of its speed and accuracy, WHAM provides a foundation for in-the-wild motion capture applications. 



\smallskip
\noindent\textbf{Limitations and future directions:}
WHAM learns about human motion from AMASS, limiting generalization to motions that are out of distribution. While we employ random masking as part of our data synthesis process, our generating approach mainly assumes the scenario where the full body is within the field of view. See Sup.~Mat.~for more details. 

WHAM opens up many directions for future work. For example, while we use SLAM to estimate the camera's angular velocity, SLAM could also provide camera intrinsics and extrinsics as well as information about the 3D scene that could be used to enforce consistency between the scene and the human. While WHAM is an online method, designed for real-time applications, it could also initialize an optimization-based post-processing akin to bundle adjustment, which would optimize the camera, scene, and human motion together. Furthermore, a real-time and phone-based implementation of WHAM should be feasible.

%% file: chapters/06_acknowledgement.tex
\smallskip
\noindent\textbf{Acknowledgements.}
\noindent 
Part of this work was done when the first author was an intern at the Max Planck Institute for Intelligence System. This work was partially supported through an NSF CAREER Award (CBET 2145473) and a Chan-Zuckerberg Essential Open Source 
Software for Science Award. 

\noindent\textbf{CoI Disclosure}.
MJB has received research gift funds from Adobe, Intel, Nvidia, Meta/Facebook, and Amazon. MJB has financial interests in Amazon and Meshcapade GmbH. While MJB is a co-founder and Chief Scientist at Meshcapade, his research in this project was performed solely at, and funded solely by, the Max Planck Society.

%% file: chapters/07_appendix.tex
As promised in the main paper, this supplemental document provides details of our synthetic data generation, the datasets we use, our network training, run-time cost, and limitations.
Additionally, please refer to our project page at \href[pdfnewwindow=true]{http://wham.is.tue.mpg.de/}{http://wham.is.tue.mpg.de/} for results that illustrate our method and recent SOTA methods applied to video sequences.

\section{Synthetic data generation}
To address the scarcity of the video data with paired 3D ground truth, we pretrain WHAM on an extensive number of synthetic 2D keypoint sequences for which we have the ground truth 3D poses. In this section, we describe the process of data synthesis from AMASS \cite{amass}.

\paragraph{2D keypoint sequence synthesis.} During training, we sample sequences of SMPL poses of length $L=81$  from AMASS. Then, similar to MEVA \cite{meva}, we uniformly upsample or downsample the frames to speed up or down the motion by up to 50\% of the original speed. 
Furthermore, we apply a random root rotation $\Delta \Gamma \sim \mathcal{U}(0^\circ, 360^\circ)$ to the axis that is vertical to the ground plane and Gaussian noise to the shape parameter $\Delta\beta \sim \mathcal{N}(0, 0.1)$. Given the augmented SMPL sequence, we extract 3D keypoints that correspond to the MS-COCO keypoints and add 3D noise modeled following previous work \cite{videonet}. Finally, we apply a random mask with the average probability of $p=0.15$ to the 3D keypoints and project them onto the virtual camera as described below.

\paragraph{Contact label generation.} The goal of generating a contact label is to train the motion decoder to detect the foot-ground contact accurately. Previous work \cite{proxycap} uses both the velocity and height of the feet to generate contact labels. However, in order to generalize our approach to arbitrary ground conditions such as slopes or stairs, we only use foot velocity to compute the ground truth contact labels, similar to TransPose \cite{transpose}. We use the heel and toe vertices of each foot to define foot-ground contact and compute the probability as below:
\[
\hat{p}^{(t)} = \frac{1}{1 + e^{\alpha (v^{(t)} - v_t)/v_t}}.
\]
\noindent We set the threshold velocity $v_t = 1\text{cm}/\text{frame}$ and the coefficient $\alpha=5$.

\paragraph{Camera motion synthesis.} We begin with generating the initial pose of the virtual camera, followed by the modeling of the camera motion. We model the initial roll and pitch angles of the camera using  Gaussian  distributions:
\begin{eqnarray*}
\gamma_r^{(0)} & \sim & \mathcal{N}(0^\circ, (5^\circ)^2),\\
\gamma_p^{(0)} & \sim & \mathcal{N}(5^\circ, (22.5^\circ)^2).
\end{eqnarray*}
\noindent Here, we do not model the initial yaw angle since it is already handled by the random SMPL root rotation $\Delta\Gamma$. 

Subsequently, we  sample the initial camera translation, using a mix of uniform and normal distributions, to capture the ground-truth 3D pose in the camera coordinates:
\begin{eqnarray*}
T_z^{(0)} & \sim & \mathcal{U}(2m, (12\rm{m})^2) - \big(R^{(0)} t^{(0)} \big)_z,\\
T_x^{(0)} & \sim & \mathcal{N}(0, 0.25^2) ~d - \big(R^{(0)} t^{(0)} \big)_x,\\
T_y^{(0)} & \sim & \mathcal{N}(0, 0.25^2) ~d - \big(R^{(0)} t^{(0)} \big)_y.
\end{eqnarray*}
\noindent Here, $d = w \cdot T_z / 2f$ is the maximum displacement of the camera to capture the 3D keypoints within the field of view, $R^{(0)}$ is the initial camera pose, $t^{(0)}$ is the initial human translation, $w$ is the image size, and $f$ is the focal length. Next, we sample the magnitude of change in the camera's extrinsics with the Gaussian distributions:
\begin{eqnarray*}
\Delta \gamma_y & \sim & \mathcal{N}(0^\circ, (45^\circ)^2),\\
\Delta \gamma_r, \Delta \gamma_p & \sim & \mathcal{N}(0^\circ, (22.5^\circ)^2),\\
\Delta T_x, \Delta T_y, \Delta T_z & \sim & \mathcal{N}(0m, (1m)^2).
\end{eqnarray*}

\noindent Finally, we interpolate the extrinsics and construct the camera's dynamic path. Here, we sample the time stamp with 20\% of noise, instead of uniform sampling, to model the non-linear camera motion. 

We use 6.7M frames in total and uniformly sample them during the synthetic data generation.





\section{Datasets}

In this section, we illustrate the datasets we use for training and testing our method.

\noindent\textbf{Human3.6M} \cite{human36m} is an indoor dataset containing individuals performing 15 distinct actions captured by both a motion-capture (mocap) system and 4 calibrated video cameras. The ground truth 3D keypoint locations are provided by the mocap. Following previous work \cite{vibe, tcmr, glot, mpsnet}, we use 5 subjects (S1, S5, S6, S7, and S8) to train our network after downsampling the mocap data to 25 fps.

\noindent\textbf{MPI-INF-3DHP} \cite{mpii3d} is a multi-view and markerless dataset containing individuals performing various ranges of motion with corresponding ground-truth 3D keypoint locations. To train the network, we use the training set of the dataset, containing 8 subjects and 16 videos per subject.

\noindent\textbf{InstaVariety} \cite{hmmr} is a large-scale in-the-wild video dataset with large variations in subjects, motion, and environment. The dataset contains pseudo-ground-truth 2D keypoints detected by OpenPose \cite{openpose}. We train our method on the training split of the dataset.

\noindent\textbf{3DPW} \cite{3dpw} is an in-the-wild video dataset containing ground truth 3D pose captured by a hand-held camera and 13 body-worn inertial sensors. We use the train, validation, and test splits of 3DPW for training, validating, and testing our method.

\noindent\textbf{RICH} \cite{rich} is a large-scale multi-view dataset captured in both indoor and outdoor environments. RICH provides the ground truth SMPL-X \cite{smplx} parameters. Following previous work \cite{bedlam}, we use the test split of the dataset to evaluate our method on 3D pose estimation accuracy.

\noindent\textbf{EMDB} \cite{emdb} is a recently captured dataset that uses a dynamic camera and body-worn electromagnetic (EM) sensors. EMDB provides ground-truth SMPL parameters as well as the global trajectory of the individuals in a global coordinate system. We use two distinct test splits, \textit{EMDB 1} and \textit{EMDB 2}, to evaluate the performance on 3D pose and shape estimation (\textit{EMDB 1}) and global trajectory estimation (\textit{EMDB 2}).

\noindent\textbf{BEDLAM} \cite{bedlam} is a recently proposed large-scale synthetic dataset. BEDLAM introduces realistic modeling of diverse clothing, hair, motion, skin tones, and scene environments to synthesize videos. BEDLAM contains 1 million video frames for individuals with ground truth SMPL/SMPL-X parameters. We optionally use the train split of BEDLAM to train the network.

\section{Losses}
The loss functions of WHAM can be categorized into two groups; motion reconstruction and trajectory reconstruction:

\begin{eqnarray*}
L_{\rm{total}} & = & \overbrace{L_{\rm{smpl}} + L_{\rm{verts}} + L_{\rm{3D}} + L_{\rm{2D}} + L_{\rm{casc}}}^\text{motion reconstruction} + \\
& & \overbrace{L_{\rm{root}} + L_{\rm{contact}} + L_{\omega} + L_{\rm{cam}} + L_{\rm{fs}}}^\text{trajectory reconstruction}.
\end{eqnarray*}

To be specific, we supervise WHAM directly on SMPL parameters and vertices when we have the labels, $\theta^*$, $\beta^*$ and $V^*$:
\begin{eqnarray*}
    L_{\rm{smpl}} &=& \sum_{t=0}^{T} \left( \lambda_\theta ||\theta^{(t)} - \theta^{*(t)}||_2 + \lambda_\beta ||\beta^{(t)} - \beta^{*(t)}||_2 \right), \\
    L_{\rm{verts}} &=& \sum_{t=0}^{T} \lambda_{\rm{verts}} ||V^{(t)} - V^{*(t)}||_1.
\end{eqnarray*}

When the 3D keypoints annotations $(x_{3D}^*)$ are available, we follow the standard MSE loss on 3D keypoints directly regressed from the Motion Encoder $(x_{3D})$ and extracted from the predicted SMPL mesh $(\hat{x}_{\rm{3D}})$:
\begin{eqnarray*}
    L_{\rm{3D}} = \sum_{t=0}^{T} \lambda_{\rm{3D}} \left(||x_{\rm{3D}}^{(t)} - x_{\rm{3D}}^{*(t)}||_2 + ||\hat{x}_{\rm{3D}}^{(t)} - x_{\rm{3D}}^{*(t)}||_2 \right).
\end{eqnarray*}

Since WHAM reconstructs full-body mesh in a cascaded manner, we use cascade loss $L_{\textrm{casc}}$ to supervise the difference between $x_{\rm{3D}}$ and $\hat{x}_{\rm{3D}}$:
\begin{eqnarray*}
    L_{\rm{casc}} = \sum_{t=0}^{T} \lambda_{\rm{casc}}||x_{\rm{3D}}^{(t)} - \hat{x}_{\rm{3D}}||_2.
\end{eqnarray*}

For 2D loss, we follow CLIFF \cite{cliff} to compute full-perspective 2D reprojection loss when the label is available, while weak-perspective 2D loss is applied for InstaVariety dataset as its full-frame image is not available:
\begin{eqnarray*}
    L_{\rm{2D}} &=& \sum_{t=0}^{T} \lambda_{\rm{2D}}||\pi \left ( \hat{x}_{\rm{3D}}^{(t)} \right )- x_{\rm{2D}}^{*(t)}||_2,
\end{eqnarray*}
where $\pi(\cdot)$ denotes the operation of perspective projection.

We supervise root trajectory and foot contact probability directly using MSE loss when the ground truth data is available:

\begin{eqnarray*}
    L_{\rm{root}} &=& \sum_{t=0}^{T} \lambda_{\Gamma}\left( || \Gamma_0^{(t)} - \Gamma^{*(t)} ||_2 + || \Gamma^{(t)} - \Gamma^{*(t)} ||_2 \right) + \\ 
    & & \sum_{t=0}^{T} \lambda_{v} \left( || v_0^{(t)} - v^{*(t)} ||_2 + || v^{(t)} - v^{*(t)} ||_2 \right), \\
    L_{\rm{contact}} &=&\sum_{t=0}^{T} \lambda_{p} || p^{(t)} - p^{*(t)} ||_2 .
\end{eqnarray*}

The Camera's pose can be expressed by the predicted root orientation in world and camera coordinates:

\begin{eqnarray*}
    R^{(t)} &=& \Gamma_c^{(t)} \left ( \Gamma^{(t)} \right )^\top ,
\end{eqnarray*}
where $\Gamma_c$ is the predicted root orientation in the camera coordinate system. As a result, we can construct novel losses based on camera pose and angular velocity:
\begin{eqnarray*}
    L_\omega &=& \sum_{t=0}^{T} \lambda_{\omega} || \omega^{(t)} - \omega^{*(t)} ||_2  , \\
    L_{\rm{cam}} &=& \sum_{t=0}^{T} \lambda_{\rm{cam}} || R^{(t)} - R^{*(t)} ||_2  ,
\end{eqnarray*}
where $\omega$ is the reconstructed camera angular velocities from $R$. 

Finally, we add foot sliding loss using the foot velocity during the contact:

\begin{eqnarray*}
    L_{\rm{fs}} &=& \sum_{t=0}^{T} \lambda_{\rm{fs}} || p^{*(t)} \odot v_{f}^{(t)} ||_2 .
\end{eqnarray*}
Here, $v_{f}^{(t)}$ is the world-coordinate foot velocity, and $\odot$ denotes the masking operation of foot contact based on the ground truth contact probability $p^{*(t)}$.

\section{Neural-network initialization}
Uni-directional RNNs introduce the challenge of differing learning objectives between the initial frames and subsequent ones due to the initialization state. Specifically, in traditional RNNs, the initialization state is typically padded with zeros, resulting in the first frame primarily relying on the input signal. In contrast, the subsequent frames are trained to capitalize on both the input signal and information transferred from the past. To resolve the disparity in learning objectives, we use a neural initialization network, as proposed by \cite{pip}, to predict $h_{0, E}$ and $h_{0, D}$ from the 0-th frame pose, instead of using zero-padding. During the training, we use pseudo-ground-truth 3D pose \cite{neuralannot} for the video datasets that do not have the SMPL parameter annotation \cite{human36m, mpii3d, hmmr}. Note that we do not supervise our network on the pseudo labels. At test time, we use the pose and shape predicted by a single-frame regressor as the initial state.

\input{figures/07_ContactExamples}
\section{Qualitative evaluation of contact estimation}
In the main paper, we show the quantitative effect of the contact-aware trajectory refinement unit. To supplement this with more intuitive visualization, we provide qualitative results in Fig. \ref{fig:contactExamples}. As shown in the figure, WHAM without the trajectory refinement unit induces significant foot sliding during the contact. Conversely, WHAM reconstructs a more feasible trajectory based on its contact estimation.

\section{Run-time cost}
While the core WHAM network presented here runs at 200fps, it relies on the input of several other methods.
Here we compute the run time of each module required by our framework on the EMDB dataset \cite{emdb}. 
The inference speed of all methods was computed on a single A100 GPU. 
We exclude running SLAM in this analysis as it can be obviated if we use gyroscope data (though real-time SLAM methods exist). 
As shown in Table \ref{tab:runtime}, our full method, with pre-processing steps, runs at around 9 fps with online inference (i.e., a batch size of 1 and no lag), and around 50 fps when run in batch mode (with resulting lag). 
We compare the core runtime of WHAM with SLAHMR \cite{slahmr}, excluding bounding box detection, person identification, and keypoints detection, for which there are real-time solutions. In this condition, WHAM takes 5 seconds (202 fps) for 1000 frames. Specifically, WHAM takes 4.3 seconds (237 fps) for image feature extraction and 0.7 seconds (1431 fps) to regress the motion and global trajectory. This is significantly faster than SLAHMR which takes 260 minutes ($<$ 0.1 fps) per 1000 frames.

\input{tables/tableX_runtime}
\input{figures/08_FailureCases}

\section{Discussions of limitation}
In Fig. \ref{fig:failureCases}, we demonstrate exemplar cases when WHAM fails to capture the global motion of the person. Since the datasets we used for training WHAM on global trajectory estimation do not contain people riding skateboards or bicycles, WHAM does not capture the global trajectory in these scenarios. Furthermore, our contact estimation only applies to the feet, thus, WHAM fails to capture ground contact of the human body other than the feet, resulting in physically infeasible body support (floating hands) or sliding of the contact points. More information on human-object interactions can be used to resolve this issue.
While we employ random masking as part of our data synthesis process, our generating approach mainly assumes the scenario where the full body is within the field of view. 
This can be addressed with additional augmentation during training (cf.~\cite{pare,osx}).

%% file: figures/07_ContactExamples.tex
\begin{figure}[ht!]
\centerline{\includegraphics[width=0.9\linewidth]{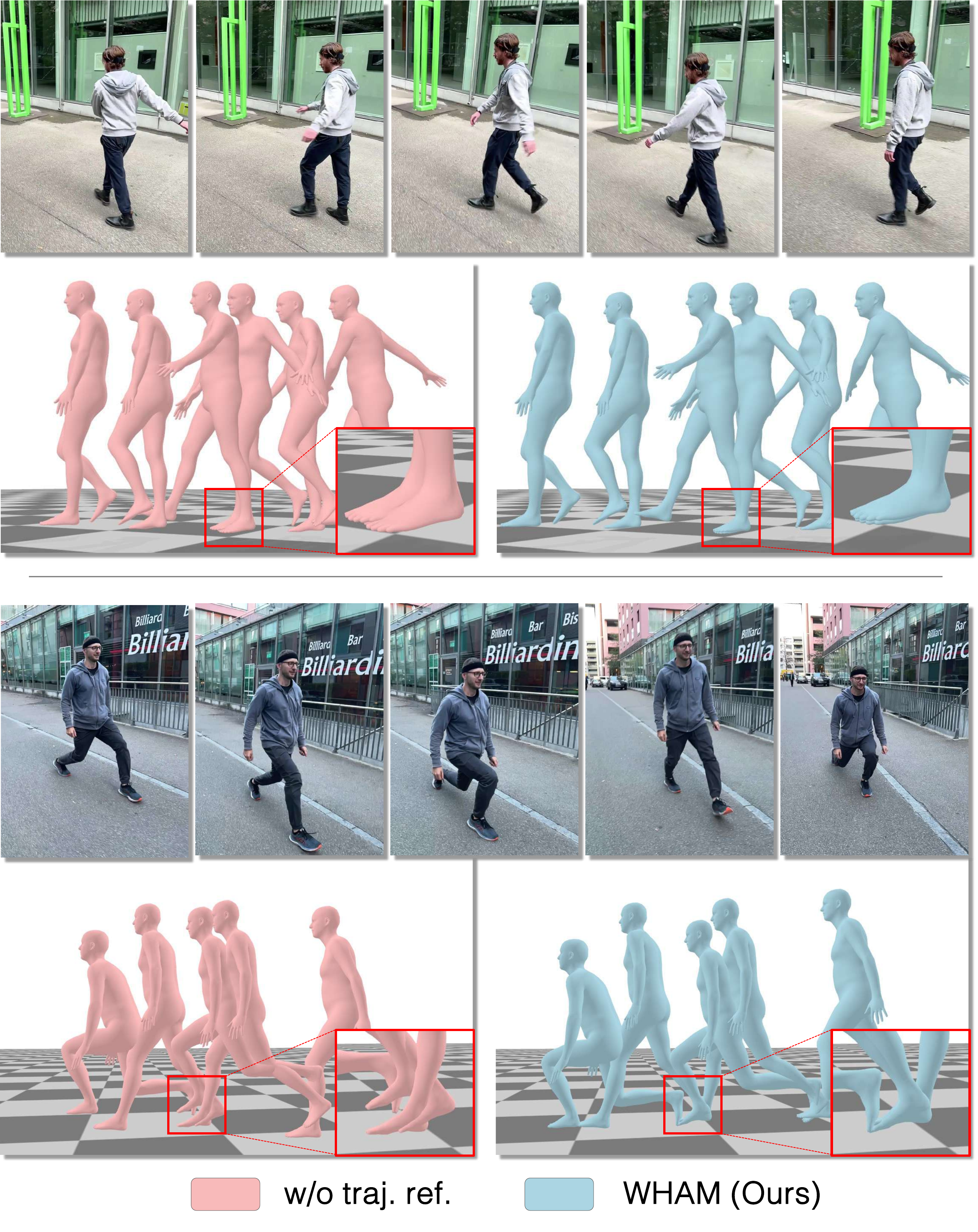}}
\vspace{-0.1in}
   \caption{Qualitative comparison between \textcolor{Turquoise}{WHAM} and after removal of contact-aware trajectory refinement (\textcolor{Salmon}{w/o~traj.~ref.}).}
\vspace*{-0.2cm}
\label{fig:contactExamples}
\end{figure}

%% file: tables/tableX_runtime.tex
\newcolumntype{?}{!{\vrule width 0.5pt}}

\begin{center}
    \begin{table}
    \setlength{\tabcolsep}{3pt}
    \resizebox{0.48\textwidth}{!}{{\small{
    \begin{tabular}{l?cc}
    \cmidrule(lr){1-3}

     & \multicolumn{2}{c}{Runtime: fps ($ms$)} \\
     \cmidrule{2-3}
    Methods & batch size = 1 & batch size = 64 \\
    \cmidrule{1-3}
    Bounding box detection & 70 (14.3 $ms$) & 265 (3.8 $ms$) \\
    Bounding box tracking & 7189 (0.1 $ms$) & 7189 (0.1 $ms$) \\
    2D keypoints detection & 12.1 (82.6 $ms$) & 88 (11.4 $ms$) \\
    Image feature extraction & 66 (15.2 $ms$) & 237 (4.3 $ms$) \\
    Rest of the framework & 926 (1.1 $ms$) & 1431 (0.7 $ms$) \\
    \cmidrule{1-3}
    Total & 8.8 (113.3 $ms$) & 49.3 (20.3 $ms$) \\
    \cmidrule{1-3}
    \end{tabular}
    }}}
    \caption{Per-frame computation time (running time) of each module in WHAM. We present this both as frames per second (fps) and milliseconds ($ms$).}
    \label{tab:runtime}
    \end{table}
\end{center}

%% file: figures/08_FailureCases.tex
\begin{figure*}[ht!]
\centerline{\includegraphics[width=0.9\linewidth]{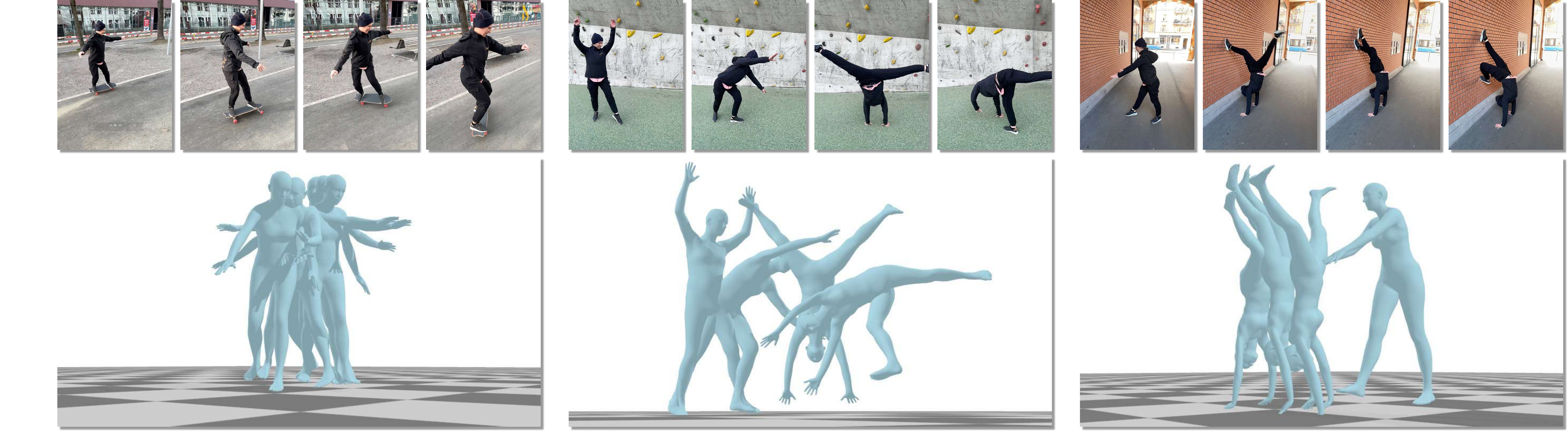}}
\vspace{-0.1in}
   \caption{Failure cases of WHAM in global motion estimation.}
\vspace*{-0.2cm}
\label{fig:failureCases}
\end{figure*}